\documentclass{article}

\usepackage[dblblindworkshop, final]{neurips_2026}

\workshoptitle{On-Device Intelligence: Foundation Models under Real-World Constraints}

\usepackage[utf8]{inputenc}
\usepackage[T1]{fontenc}
\usepackage{hyperref}
\usepackage{url}
\usepackage{booktabs}
\usepackage{amsfonts}
\usepackage{amsmath}
\usepackage{amssymb}
\usepackage{nicefrac}
\usepackage{microtype}
\usepackage{graphicx}
\usepackage{multirow}
\usepackage{xcolor}

\usepackage{xcolor}
\newcommand{\tb}[1]{\textcolor{black}{#1}}

\title{DCM-SAM: Defect-Conditioned Mixture of LoRA Experts for NPU-Deployed AM Defect Segmentation}

\author{%
  Md Mushfiqur Rahaman$^{2,3,*}$ \hspace{0.4em} Md Mahedi Hasan$^{1,*}$ \hspace{0.4em} Imtiaz Ahmed$^{2}$ \hspace{0.4em} Srinjoy Das$^{1,2,3,\dagger}$ \\
  $^{1}$Lane Department of Computer Science \& Electrical Engineering \\
  $^{2}$Department of Industrial and Management Systems Engineering \\
  $^{3}$School of Mathematical and Data Sciences \\
  West Virginia University, Morgantown, WV \\
  \texttt{\{mr00131, mh00062\}@mix.wvu.edu} \\
  \texttt{\{imtiaz.ahmed, srinjoy.das\}@mail.wvu.edu} \\
  $^{*}$Equal contribution. \quad $^{\dagger}$Corresponding author: \texttt{srinjoy.das@mail.wvu.edu}
}

\begin{document}

\maketitle

\begin{abstract}
\tb{Metal additive manufacturing parts are inspected by X-ray computed tomography, where labelled data is scarce, the pores and inclusions that matter span a few pixels, and inspection must happen at the machine. We present DCM-SAM, a defect-conditioned adaptive mixture of LoRA experts: one frozen Segment Anything backbone carries a separate Conv-LoRA expert bank and mask decoder per defect class, each trained in its own pass, without prompts, on synthetic slices alone, updating only $4.4\%$ of the parameters. On benchmarks that XCT-SAM reports, DCM-SAM improves on every baseline for both classes from a ViT-B backbone against their ViT-H, and reaches $64.2\%$ pore IoU on real NIST scans having seen no real images during training. Deployment then exposes what adaptation work rarely measures: on a Qualcomm Hexagon NPU, ViT-H and ViT-L compile yet cannot allocate at $1024^2$ image resolution, since activations rather than weights exceed the device ceiling, and quantizing weights does not help. ViT-B alone runs, but the adapted encoder then fails to allocate where the stock one succeeds, until a numerically identical rewrite of the attention lets the complete DCM-SAM run in FP16 at $1024^2$, with no operator falling back to the CPU, masks within $0.01\%$ of pixels of the FP32 reference. Code: \href{https://github.com/MushfiqShovon/DCM-SAM}{https://github.com/MushfiqShovon/DCM-SAM}.}
\end{abstract}
\section{Introduction}
\label{sec:intro}
\tb{In metal additive manufacturing (AM) where parts are built up layer by layer, e.g. via laser powder bed fusion, small fluctuations in laser power or melt-pool behaviour leave behind porosity, lack-of-fusion voids and foreign inclusions, and these decide fatigue life and whether a part can be certified~\cite{ngo2018additive,wang2026recent,tusher2025comprehensive,haribaskar2024defects,wang2026porosity,zhan2025assessment}}.
Because the defects are small, sparse and buried below the
surface~\cite{poudel2022feature,lapre2024rapid,bimrose2025detecting}, \tb{X-ray computed tomography (XCT)-based} inspection has become the
reference way of finding them~\cite{sun2025x,jones2025validation,baig2025non}, and the
bottleneck is now \tb{accurately} interpreting the scans. Classical threshold pipelines need re-tuning for
every dataset~\cite{ouidadi2023defect, perghem2025ml,ledwaba2025development}, and supervised \tb{segmentation} networks carry the statistics of their training data~\cite{unet++}. 
\tb{Three properties of the task shape this paper: the defects that matter are a few pixels wide even after a slice is upsampled to the model's $1024^2$ input, annotated real volumes are rare, and a model trained on one alloy and scanner is routinely asked to work on another. Two consequences follow. First, the accuracy the task needs comes from adapting a segmentation foundation model such as SAM~\cite{sam_2023} rather than training a small one from scratch: a UNet++ baseline reaches only $13.4\%$ pore IoU on real NIST scans (Table~\ref{tab:baselines}). Second, the defect scale makes full resolution a hard floor rather than a preference: the same weights evaluated at $512^2$ lose almost half their pore IoU, and retraining at $512^2$ does not recover it (\S\ref{sec:exp:deploy}). A third constraint comes not from the data but from where the model is allowed to run: cloud offload is ruled out by data locality, since AM inspection data is frequently proprietary or export-controlled, and a workstation per machine does not scale across a fleet at comparable cost and power, so the target is an embedded accelerator. AM XCT therefore forces a foundation-model-scale backbone and full input resolution onto an embedded device at once, and that combination is what surfaces the constraints reported in \S\ref{sec:exp:backbone}--\S\ref{sec:exp:deploy}.}

\paragraph{Prior work.} The Segment Anything Model (SAM)~\cite{sam_2023} and its
descendants~\cite{sam2_2024,medsam_2024,sam_med2d} bring segmentation priors no industrial dataset could provide, and several groups have transferred them to AM
XCT~\cite{era2025unsupervised,tabassum_2024_adapting,gruber2024adapting,ma2025alloy}. Fully fine-tuning \tb{SAM's 637M-parameter encoder} on a few hundred slices overfits, so the usual
approach freezes it and learns small injected modules: LoRA~\cite{lora_2022} learns
low-rank weight updates, and Conv-LoRA~\cite{conv_lora_2024} adds multi-scale convolutions inside the bottleneck, routed by a sparsely gated mixture of
experts~\cite{shazeer2017outrageously}; Tabassum and Ziabari~\cite{tabassum_2024_adapting}
applied it to AM XCT with CycleGAN-synthesised supervision, and XCT-SAM~\cite{xctsam_2026}
added an intermediate adaptation stage on alloy microstructure.
\tb{Both train a separate model per class but never ask what running it costs;} post-training
quantization~\cite{ptq4sam_2024,brecq_2021,qdrop_2022,adaround_2020} reports analytical rather than measured speedups, and distilled
backbones~\cite{mobilesam_2023,efficientvit_sam_2024} give up the capacity \tb{that} adaptation was meant to provide.

\paragraph{This work.} Pores are common but only a few pixels wide, whereas inclusions are
an order of magnitude rarer and occupy under $1\%$ of a slice. A single low-rank subspace
has to serve both, and because pores vastly outnumber inclusions its updates are dominated
by pore examples, leaving inclusion learning noisy and prone to overfitting. Fully separate
models avoid this interference but give up the frozen encoder they could share. \textbf{DCM-SAM} (\emph{Defect-Conditioned Adaptive Mixture of LoRA Experts}) takes
the middle path: one frozen SAM backbone hosts a separate Conv-LoRA expert bank and mask
decoder for each defect class, each trained in its own pass with its own loss and sampling,
so the heads share no trainable parameters and cannot interfere, and both run without
prompts. The expert count is chosen per class; a top-$1$ gate then selects one expert per
input. 
\tb{We deploy this design on SAM ViT-B which is the only one of SAM's three backbones whose
attention activations, not its weights, fit our Qualcomm Dragonwing IQ-9075 NPU target at
$1024^2$ (\S\ref{sec:exp:backbone}) and run the selected $(2,8)$ expert configuration
end to end (\S\ref{sec:exp:deploy}). Our contributions are as follows:}

\begin{itemize}\setlength{\itemsep}{1pt}\setlength{\topsep}{2pt}
  \item \textbf{On-device deployment.} We measure which SAM backbones actually execute on \tb{the} NPU and explain why the others do not, and we deploy the complete adapted model at $1024^2$ \tb{image resolution} with every operator on the NPU, producing masks that differ from FP32 reference in under $0.01\%$ of pixels (\S\ref{sec:exp:backbone}, \S\ref{sec:exp:deploy}). \tb{The following findings reach beyond AM.} Whether a backbone fits is decided by its activations rather than its weights, and quantization cannot move that ceiling at the bit-widths this toolchain supports. An adapter holding only $0.08\%$ of the parameters can still decide whether a graph fits. Top-$1$ routing, which is free on a GPU, costs dense compute once the graph is made static, $4.1\times$ from $M{=}2$ to $M{=}8$. \tb{The best expert count transfers neither across backbones nor from synthetic to real data.} 
  \item \textbf{Domain adaptation.} We place per-class Conv-LoRA expert banks and decoders over one frozen SAM and train them without prompts. Each ViT-B head trains $4.13$M parameters, $4.4\%$ of the model ($8.27$M and $8.4\%$ for both heads); the adapter itself is $0.08\%$, and the rest is a mask decoder trained from scratch. \tb{On the baselines XCT-SAM~\cite{xctsam_2026} reports, the model improves on every one, subject to the protocol caveat in \S\ref{sec:exp:ood}.}
  \item \textbf{Data-efficient learning.} \tb{The model is adapted from a few hundred CycleGAN-synthesised slices with no intermediate alloy-microstructure stage like XCT-SAM's, and never sees a real image; yet what it learns carries over to real scans (\S\ref{sec:exp:ood}).}
  \item \tb{\textbf{Architecture selection under shift.}} On real NIST XCT, most of the differences that the synthetic ablation \tb{reveals} between configurations disappear under distribution
  shift (\S\ref{sec:exp:ood}).
\end{itemize}

\section{Method}
\label{sec:method}

\begin{figure}[t]
\centering
\includegraphics[width=0.85\textwidth]{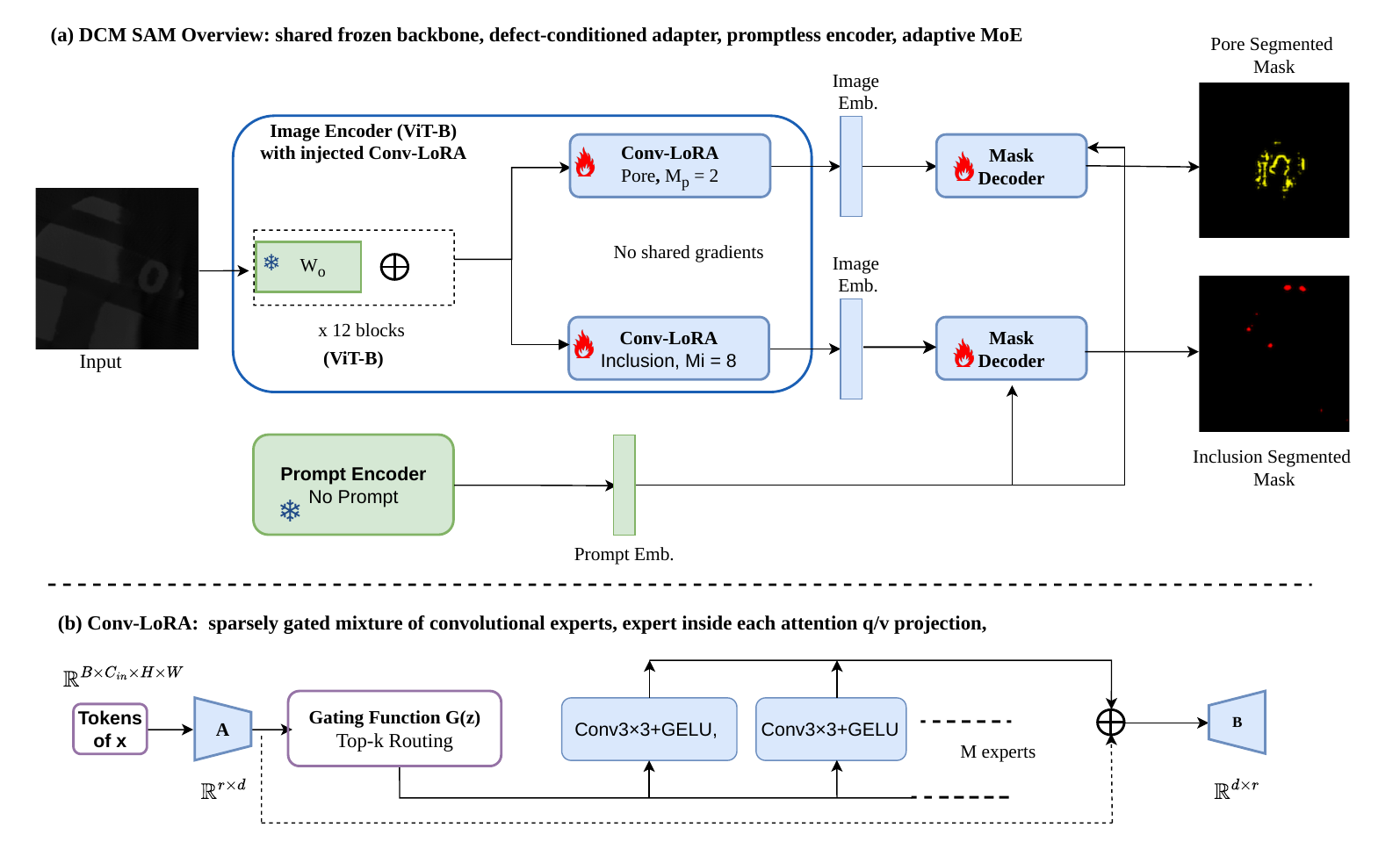}
\caption{DCM-SAM. (a) Conv-LoRA with a mixture of experts is injected into every block of one
frozen SAM ViT-B, giving each defect class its own encoder path and decoder ($M_p=2$ experts
for pores, $M_i=8$ for inclusions); the heads share the frozen weights but no trainable
parameter, and the encoder runs once per head. (b) Within a block, a top-$k$ gate routes the
low-rank bottleneck to one of $M$ convolutional experts, each at its own spatial scale.}
\label{fig:arch}
\vspace{-3mm}
\end{figure}

\paragraph{Promptless SAM.} SAM~\cite{sam_2023} has three parts: a ViT image encoder
$f_{\text{enc}}$, a prompt encoder $f_{\text{prm}}$ and a light mask decoder $f_{\text{dec}}$.
Given an image $x \in \mathbb{R}^{3\times S\times S}$ with $S=1024$, the encoder splits it
into $N=(S/16)^2=4096$ patch tokens and passes them through $L$ transformer blocks, four with global attention and the rest attending within $14\times14$ windows, producing an embedding $E \in \mathbb{R}^{256\times\frac{S}{16}\times\frac{S}{16}}$. The encoder contains $96\%$ of the parameters for ViT-B ($99\%$ for ViT-H) and is evaluated once per image. Since industrial inspection may involve hundreds of defects per slice, explicit prompting is impractical. We therefore retain $f_{\text{prm}}$ without providing any prompts; instead, it produces only its learned ``no-prompt'' embeddings. Defect localization is thus driven entirely by the learned image embeddings and a $4.1$M-parameter mask decoder trained from scratch.

\paragraph{Conv-LoRA with a multi-scale expert mixture.} LoRA~\cite{lora_2022} adapts a frozen projection $W_0$ as $h=W_0x+\frac{\alpha}{r}BAx$ using two low-rank matrices of rank $r$. This linear update treats tokens independently and cannot capture the local structure needed for defects spanning only a few pixels. Conv-LoRA~\cite{conv_lora_2024} addresses this limitation by injecting local spatial structure into the bottleneck. Specifically, we reshape $Z=Ax$ into the $\frac{S}{16}\times\frac{S}{16}$ patch grid, select one of $M$ convolutional experts, and add its output to $Z$ before projecting back to the original dimension.
\begin{equation}
h = W_0x + \tfrac{\alpha}{r}\,B\Big[Z + \mathcal{E}_{i^\star}(Z)\Big],
\qquad
\mathcal{E}_i(Z) = \mathcal{D}_i\big(\sigma(\mathrm{Conv}_{3\times3}(\mathcal{U}_i(Z)))\big),
\label{eq:convlora}
\end{equation}
where $\mathcal{U}_i$ upsamples bicubically by a factor $i$, $\mathcal{D}_i$ resamples back, and $\sigma$ is a GELU. A $3\times3$ kernel on a feature map upsampled by $i$ covers a region $i$ times smaller, giving the experts a ladder of receptive fields whose depth is set by the class-specific expert count (\S\ref{sec:exp:ablation}). This multi-scale coverage is important even within a class, as pore components in a single slice span a $40\times$ range in area. 

A load-balancing loss, $\mathcal{L}_{\text{moe}}=\mathrm{CV}^2(\text{importance})+\mathrm{CV}^2(\text{load})$, encourages all experts to remain active~\cite{shazeer2017outrageously}. Only one expert executes per input on a GPU, although this sparsity is lost on a static NPU graph (\S\ref{sec:exp:deploy}). We use $r=2$ and $\alpha=8$, inserting one adapter into each encoder block.

\paragraph{Per-class heads with gradient isolation.} DCM-SAM, illustrated in Figure~\ref{fig:arch}, uses a separate head for each class $c\in{\{\text{pore},\text{inclusion}\}}$, with class-specific adapters $\theta_c^{\text{lora}}$ and decoder $\theta_c^{\text{dec}}$ built on a shared, frozen backbone $\theta^{\text{sam}}$. The heads are optimized in parallel but independently, using separate optimizers, learning-rate schedules, and gradient clipping to prevent cross-head interference. As a result, they produce distinct embeddings for the same input image (cosine similarity $0.87$). At inference, the encoder runs once per head, and each deployed binary contains its own copy of the frozen backbone weights (\S\ref{sec:exp:deploy}; see Appendix~\ref{app:limits}). Each
head trains $4.13$M parameters, $4.4\%$ of the ViT-B model (Appendix~\ref{app:params}), using
$\mathcal{L}_c=\mathcal{L}_{\text{dice}}+\mathcal{L}_{\text{focal}}+0.1\,\mathcal{L}_{\text{moe}}$. Inclusion-positive images are oversampled fivefold, while the inclusion loss is averaged per image to prevent densely labeled slices from dominating individual training steps. See Appendix~\ref{app:train}.

\paragraph{Deployment.} Three things have to be settled before the model can run on an NPU.
The first is memory. Each global attention block forms a matrix with one entry per pair of
tokens, and the token count $N=(S/16)^2$ grows with the square of the input side $S$, so with
$H$ heads the attention activations grow as $HN^2$, the fourth power of $S$. Halving the input
side cuts the attention matrix sixteen-fold; halving the precision only halves it. That is
why resolution, not precision, is the lever in \S\ref{sec:exp:backbone}, whose $512^2$ runs
use a retargeted model (Appendix~\ref{app:npu}). The second is routing. The gate chooses an
expert by an $\arg\max$ on its input, which a static graph cannot express, so for export we
compute all $M$ experts and select one with a one-hot mask; over eight slices this changes \tb{only}
$13$ of $8.4$M mask pixels (Appendix~\ref{app:npu}). The third is how attention is written.
SAM's reference code builds the attention matrix through several full-size intermediates that
a static graph must hold at once; the same computation as one scaled-dot-product attention
call with a single additive bias agrees to $1.2\times10^{-5}$ and keeps far fewer tensors
alive, which \S\ref{sec:exp:deploy} shows is what lets the adapted model fit
(Appendix~\ref{app:attn}). \tb{Finally, the device is rated for INT8, yet we deploy in FP16 and report no INT8 numbers; \S\ref{sec:exp:backbone} explains why.}
\vspace{-1.5mm}
\section{Experiments}
\label{sec:experiments}
\vspace{-1.5mm}
\paragraph{Setup.} We train on CycleGAN-synthesized AM XCT slices and evaluate on five held-out synthetic splits of $100$ images each, reporting mean IoU and Dice across splits. We also assess out-of-distribution performance on real NIST AM XCT scans. All checkpoints use the same fixed per-split post-processing (see Appendix~\ref{app:train}). Both heads are trained with AdamW, layer-wise decay, warmup followed by cosine decay for $30$ epochs. IoU and Dice are averaged over images, with defect-only results reported in Appendix~\ref{app:train}. We identified and removed a $4.9$-point pore IoU seeding confound before reporting any results (Appendix~\ref{app:seed}). Since each configuration uses a single seed, differences smaller than roughly three points should be interpreted carefully. Our target edge device is the Qualcomm Dragonwing IQ-9075, equipped with a Hexagon v73 NPU rated at $100$ INT8 TOPS; we collected all NPU measurements reported in this study directly on the device.  
\vspace{-1.5mm}
\subsection{How many mixture-of-experts?}
\label{sec:exp:ablation}
\vspace{-1.5mm}
We sweep one head's mixture-of-experts while fixing the other at $M=8$, in both directions, and directly train the corner and cross configurations.
\tb{On ViT-H, IoU is flat above $M=2$, so the study selects $(M_{pore}, M_{incl})=(2,1)$ (Appendix~\ref{app:vith}).}
Table~\ref{tab:vitb} reports the nine ViT-B configurations we evaluated. Three findings stand out. First, XCT-SAM's default $M=8$ for both heads performs worst, achieving only $25.8$ pore IoU. This degradation is specific to the joint $(8,8)$ setting rather than to $M_{\text{pore}}=8$: $(8,1)$ achieves the best pore IoU. Second, the per-head optima do not combine directly. Although each independent sweep selects $M=2$, the joint $(2,2)$ configuration underperforms $(2,8)$ on both metrics, showing that joint configurations must be optimized explicitly. Third, the optimal mixture size is backbone-dependent: ViT-H remains relatively stable for $M>2$ and favors a smaller pairing, whereas ViT-B degrades sharply when both mixture sizes are large and performs poorly with $(2,2)$. In this study, we use $(2,8)$ as the best-balanced configuration, achieving a summed IoU of $85.0$. The $(8,1)$, $(8,2)$, and $(8,4)$ configurations are all within $1.0$ point of this result, a difference that cannot be resolved with a single seed. We therefore treat $(2,8)$ as a deployment choice rather than a claim that it is the globally optimal mixture size.

\begin{table}[t]
\centering\scriptsize
\setlength{\tabcolsep}{5pt}
\renewcommand{\arraystretch}{0.92}
\caption{ViT-B configurations: mean IoU (\%) over the five synthetic splits, and pore IoU on real NIST XCT. Bold marks the column maximum.}
\label{tab:vitb}
\begin{tabular}{lccc @{\hspace{1.5em}} c}
\toprule
$(M_{\text{pore}}, M_{\text{incl}})$ & Role & Synth.\ pore & Synth.\ incl. & \textbf{NIST pore (OOD)} \\
\midrule
$(1,1)$ & corner            & 32.8 & \textbf{49.6} & 64.4 \\
$(1,8)$ & pore sweep        & 29.4 & 47.7 & 65.1 \\
$(2,2)$ & composed optimum  & 35.0 & 46.3 & 64.3 \\
$(2,8)$ \emph{(selected)} & pore sweep & 36.1 & 48.9 & 64.2 \\
$(4,8)$ & pore sweep        & 34.6 & 44.9 & 65.8 \\
$(8,1)$ & incl.\ sweep      & \textbf{36.5} & 47.5 & 65.3 \\
$(8,2)$ & incl.\ sweep      & 34.6 & 49.5 & \textbf{66.3} \\
$(8,4)$ & incl.\ sweep      & 35.5 & 48.9 & \textbf{66.3} \\
$(8,8)$ & default           & 25.8 & 46.6 & 65.3 \\
\bottomrule
\end{tabular}
\vspace{-4mm}
\end{table}

\vspace{-1.5mm}
\subsection{Out-of-distribution evaluation on real NIST XCT}
\label{sec:exp:ood}
In addition to the synthetic benchmark, we also evaluate on real NIST scans acquired using a different material and scanner. Two observations emerge from the last column of Table~\ref{tab:vitb}. First, all configurations transfer well, achieving around $30$ points higher pore IoU on NIST than on the synthetic data, likely because pores are more clearly resolved in the real scans. The selected model reaches $64.2$, or $58.8$ without \texttt{test6}, whose post-processing is specimen-specific; its errors are primarily due to over-prediction rather than missed pores (Appendix~\ref{app:nist}). Second, the ablation spread largely disappears: the $10.7$-point synthetic gap shrinks to $2.1$ points on NIST, below the noise floor reported in \S\ref{sec:experiments}.   Against the XCT-SAM~\cite{xctsam_2026} baselines (Table~\ref{tab:baselines}), DCM-SAM outperforms all three settings, with the largest gains on inclusions, where Conv-LoRA-SAM degrades despite competitive pore performance. However, these baselines employ ViT-H, whereas DCM-SAM uses ViT-B.

\begin{table}[t]
\centering\scriptsize
\renewcommand{\arraystretch}{0.92}
\caption{Comparison against the baselines reported by XCT-SAM~\cite{xctsam_2026} (mean IoU, \%).
DCM-SAM uses ViT-B with $(2,8)$; all baselines use ViT-H.}
\label{tab:baselines}
\begin{tabular}{lccc}
\toprule
Method & Pores (synthetic) & Inclusions (synthetic) & Pores (NIST, OOD) \\
\midrule
SAM~\cite{sam_2023} \emph{(zero-shot)} & 14.3 & 29.5 & 49.4 \\
UNet++~\cite{unet++}                   & 14.0 & 10.2 & 13.4 \\
MedSAM~\cite{medsam_2024}              & 16.3 & 32.5 & 33.8 \\
SAM-Med2D~\cite{sam_med2d}             & 18.7 & 36.0 & 57.8 \\
Conv-LoRA-SAM~\cite{conv_lora_2024}    & 27.7 & 24.1 & 54.6 \\
XCT-SAM~\cite{xctsam_2026}             & 32.5 & 36.4 & 58.5 \\
\midrule
\textbf{DCM-SAM (ours)}                & \textbf{36.1} & \textbf{48.9} & \textbf{64.2} \\
\bottomrule
\end{tabular}
\vspace{-4mm}
\end{table}

\vspace{-1.5mm}
\subsection{Backbone selection under edge constraints}
\label{sec:exp:backbone}
\vspace{-1.5mm}
We compiled all three SAM encoders ($89.7$M, $308.3$M and $637$M parameters) for the Hexagon v73 and tried to execute them at $1024^2$; all three compile, only ViT-B runs. What limits the others is activation memory, not weights. The same ViT-H weights run at $512^2$ and fail at $1024^2$, ViT-L still cannot allocate with INT8 weights, and at $1024^2$ one global-attention layer materialises a $16\times4096\times4096$ tensor ($512$\,MB in FP16) that the backend has no fused kernel to avoid. Hidden size matters through the head count and the weights rather than the per-token vectors, and the roughly $8$\,MB of on-chip vector memory on this Hexagon generation~\cite{qnn_vtcm_sharing,lam2024hexagon} is exceeded by every backbone and simply tiled through \tb{ViT-B included, but is tiled through without consequence} (Appendix~\ref{app:attn}). Quantization does not move this ceiling, because the attention products $QK^{\top}$ and $AV$ take two \emph{activation} operands and the backend has no validated integer kernel for that pairing at $8$ or $16$ bits; resolution does, since at $512^2$ both ViT-L and ViT-H run ($4.14$\,s and $6.21$\,s per image). \tb{Token count, not bit-width, is therefore the lever for feasible inference on the edge device.}
\vspace{-1.5mm}
\subsection{Deploying the full DCM-SAM on the NPU}
\label{sec:exp:deploy}
\vspace{-1.5mm}
The selected ViT-B $(2,8)$ model deploys as four compiled graphs, an encoder and a decoder per head, since each head carries its own adapters inside the encoder. Exported as trained, the adapted ViT-B encoder needs $4.07$\,GB at $1024^2$ against a $3.76$\,GB ceiling set by
the graph serializer, although the stock encoder with the same eager attention runs; only
the two together exceed it (Appendix~\ref{app:attn}).
\tb{With the scaled-dot-product formulation of \S\ref{sec:method}, all four graphs execute at $1024^2$ with no operator falling back to the CPU. On one real slice per head, the device masks differ end to end in comparison to the FP32 reference in only $96$ (pore) and $0$ (inclusion) of a total of $1{,}048{,}576$ pixels, this is the model that Table~\ref{tab:attn} in Appendix~\ref{app:attn} reports.}

Three findings follow. First, static routing makes the mixture size an inference cost that scales steeply with $M$: the two encoders differ only in $M$, yet the $M=8$ head is $4.1$ times slower, because a GPU discards the unselected experts at no measurable cost while the NPU pays for every one of them (Appendix~\ref{app:npu}). So $(1,1)$, the cheapest configuration, also holds the best inclusion IoU in the study, for $3.3$ pore points, within the $4.9$-point seeding swing of Appendix~\ref{app:seed}. Second, the decoder is essentially free, $500$ times cheaper than the encoder: defect-awareness costs one extra encoder pass and nothing else. 
\tb{Third, resolution cannot be traded for speed, even with retraining: a ViT-H head retrained at $512^2$ still ends $5.6$ points below its $1024^2$ counterpart on pores ($12$ to $50\%$ per split on defect-bearing images), while inclusion holds; without retraining the same weights lose almost half their pore IoU ($36.1$ to $19.7$).}
The two ways of fitting the device thus trade different classes, resolution costing pores and backbone costing inclusions, and we chose the side that keeps pores, \tb{the only class real data lets us score} (Table~\ref{tab:res512} in Appendix~\ref{app:train}).
\vspace{-1.5mm}
\section{Conclusion}
\label{sec:conclusion}
\vspace{-1.5mm}

\tb{DCM-SAM adapts a frozen SAM ViT-B to two distinct defect classes while training only $4.4\%$ of the parameters per head. Using a few hundred synthetic slices, it achieves $64.2\%$ pore IoU on real NIST scans, exceeding XCT-SAM's reported results despite using SAM backbones once. It runs at full resolution on the IQ-9075 with every operator on the NPU, masks within $0.01\%$ of the FP32 reference which is evidence that adapting, not shrinking or quantizing, a frozen foundation model is what meets edge AM inspection's accuracy and deployment constraints together.}

\begin{ack}
This work was supported by a gift from Qualcomm Incorporated. The authors would also like to acknowledge the Pacific Research Platform, NSF Project ACI-1541349, and Larry Smarr (PI, Calit2 at UCSD) for providing the computing infrastructure used in some of the experiments.
\end{ack}

\bibliographystyle{plain}
\bibliography{main}

@inproceedings{focal_tversky_2019,
  title={A novel focal tversky loss function with improved attention u-net for lesion segmentation},
  author={Abraham, Nabila and Khan, Naimul Mefraz},
  booktitle={2019 IEEE 16th international symposium on biomedical imaging (ISBI 2019)},
  pages={683--687},
  year={2019},
}

@article{unet++,
  title={U{N}et++: Redesigning skip connections to exploit multiscale features in image segmentation},
  author={Zhou, Zongwei and Siddiquee, Md Mahfuzur Rahman and Tajbakhsh, Nima and Liang, Jianming},
  journal={IEEE transactions on medical imaging},
  volume={39},
  number={6},
  pages={1856--1867},
  year={2020},
  publisher={ieee}
}

@article{tabassum_2024_adapting,
  title={Adapting segment anything model (SAM) to experimental datasets via fine-tuning on GAN-based simulation: A case study in additive manufacturing},
  author={Tabassum, Anika and Ziabari, Amirkoushyar},
  journal={arXiv preprint arXiv:2412.11381},
  year={2024}
}

@inproceedings{sam_2023,
  title={Segment anything},
  author={Kirillov, Alexander and Mintun, Eric and Ravi, Nikhila and Mao, Hanzi and Rolland, Chloe and Gustafson, Laura and Xiao, Tete and Whitehead, Spencer and Berg, Alexander C and Lo, Wan-Yen and others},
  booktitle={IEEE/CVF International Conference on Computer Vision},
  pages={4015--4026},
  year={2023}
}

@inproceedings{conv_lora_2024,
  title={Convolution Meets LoRA: Parameter Efficient Finetuning for Segment Anything Model},
  author={Zhong, Zihan and Tang, Zhiqiang and He, Tong and Fang, Haoyang and Yuan, Chun},
  booktitle={International Conference on Learning Representations},
  year={2024}
}

@article{medsam_2024,
  title={Segment anything in medical images},
  author={Ma, Jun and He, Yuting and Li, Feifei and Han, Lin and You, Chenyu and Wang, Bo},
  journal={Nature communications},
  volume={15},
  number={1},
  pages={654},
  year={2024},
}

@article{sam_med2d,
  title={{SAM-Med2D}},
  author={Cheng, Junlong and Ye, Jin and Deng, Zhongying and Chen, Jianpin and Li, Tianbin and Wang, Haoyu and others},
  journal={arXiv preprint arXiv:2308.16184},
  year={2023}
}

@article{sam2_2024,
  title={SAM 2: Segment Anything in Images and Videos},
  author={Ravi, Nikhila and Gabeur, Valentin and Hu, Yuan-Ting and Hu, Ronghang and Ryali, Chaitanya and Ma, Tengyu and others},
  journal={arXiv preprint arXiv:2408.00714},
  year={2024}
}

@article{shazeer2017outrageously,
  title={Outrageously large neural networks: The sparsely-gated mixture-of-experts layer},
  author={Shazeer, Noam and Mirhoseini, Azalia and Maziarz, Krzysztof and Davis, Andy and Le, Quoc and Hinton, Geoffrey and Dean, Jeff},
  journal={arXiv preprint arXiv:1701.06538},
  year={2017}
}

@article{ngo2018additive,
  title={Additive manufacturing (3D printing): A review of materials, methods, applications and challenges},
  author={Ngo, Tuan D and Kashani, Alireza and Imbalzano, Gabriele and Nguyen, Kate TQ and Hui, David},
  journal={Composites Part B: Engineering},
  volume={143},
  pages={172--196},
  year={2018},
  publisher={Elsevier}
}

@article{lapre2024rapid,
  title={Rapid non-destructive inspection of sub-surface defects in 3{D} printed alumina through 30 layers with 7 $\mu$m depth resolution},
  author={Lapre, C and Brouczek, D and Schwentenwein, M and Neumann, K and Benson, N and Petersen, CR and Bang, O and Israelsen, NM},
  journal={Open Ceramics},
  volume={18},
  pages={100611},
  year={2024},
  publisher={Elsevier}
}

@article{bimrose2025detecting,
  title={Detecting and classifying hidden defects in additively manufactured parts using deep learning and X-ray computed tomography},
  author={Bimrose, Miles V and Hu, Tianxiang and McGregor, Davis J and Wang, Jiongxin and Tawfick, Sameh and Shao, Chenhui and Liu, Zuozhu and King, William P},
  journal={Journal of Intelligent Manufacturing},
  volume={36},
  number={5},
  pages={3465--3479},
  year={2025},
  publisher={Springer}
}

@article{sun2025x,
  title={X-ray computed tomography in metal additive manufacturing: A review on prevention, diagnostic, and prediction of failure},
  author={Sun, X and Huang, L and Xiao, BG and Zhang, Q and Li, JQ and Ding, YH and others},
  journal={Thin-Walled Structures},
  volume={207},
  pages={112736},
  year={2025},
  publisher={Elsevier}
}

@article{jones2025validation,
  title={Validation of X-ray computed tomography detection limits for stochastic flaws in additively manufactured Ti-6Al-4 V},
  author={Jones, Griffin and Sundar, Veeraraghavan and Reed, Rachel and Stecko, Marissa and Keist, Jayme},
  journal={Journal of Materials Engineering and Performance},
  volume={34},
  number={10},
  pages={8683--8690},
  year={2025},
  publisher={Springer}
}

@article{baig2025non,
  title={Non-destructive detection of critical defects in additive manufacturing},
  author={Baig, Shaharyar and Jam, Alireza and Beretta, Stefano and Shao, Shuai and Shamsaei, Nima},
  journal={Scientific Reports},
  volume={15},
  number={1},
  pages={6740},
  year={2025},
  publisher={Nature Publishing Group UK London}
}

@article{era2025unsupervised,
  title={An unsupervised approach towards promptable porosity segmentation In laser powder bed fusion by segment anything},
  author={Era, Israt Zarin and Ahmed, Imtiaz and Liu, Zhichao and Das, Srinjoy},
  journal={npj Advanced Manufacturing},
  volume={2},
  number={1},
  pages={10},
  year={2025},
  publisher={Nature Publishing Group UK London}
}

@article{perghem2025ml,
  title={Ml-based detection of critical defects in additively manufactured parts via X-ray computed tomography},
  author={Perghem, Daniel and Salehnasab, Behnam and Beretta, Stefano and Shao, Shuai and Shamsaei, Nima},
  journal={Materials \& Design},
  pages={115184},
  year={2025},
  publisher={Elsevier}
}

@article{gruber2024adapting,
  title={Adapting the segment anything model for volumetric x-ray data-sets of arbitrary sizes},
  author={Gruber, Roland and R{\"u}ger, Steffen and Wittenberg, Thomas},
  journal={Applied Sciences},
  volume={14},
  number={8},
  pages={3391},
  year={2024},
  publisher={MDPI}
}

@article{ma2025alloy,
  title={Alloy microstructure segmentation through SAM and domain knowledge without extra training},
  author={Ma, Xudong and Zhang, Yuqi and Wang, Chenchong and Xu, Wei},
  journal={Scripta Materialia},
  volume={260},
  pages={116581},
  year={2025},
  publisher={Elsevier}
}

@article{wang2026recent,
  title={Recent Advances in Metal Additive Manufacturing: Materials Design and Artificial Intelligence Applications},
  author={Wang, Shuo and Zhou, Lin and Zhong, Shiyu and Li, Gan and Zhang, Lei and Wang, Xu and Li, Zhiqiang and Lu, Jian},
  journal={Engineering},
  year={2026},
  publisher={Elsevier}
}

@article{tusher2025comprehensive,
  title={Comprehensive review of fabrication process parameters influencing defect formation in laser powder bed fused (L-PBF) Al-Si alloys},
  author={Tusher, Md Mehide Hasan and Ince, Ayhan},
  journal={Materials \& Design},
  pages={114374},
  year={2025},
  publisher={Elsevier}
}

@article{haribaskar2024defects,
  title={Defects in metal additive manufacturing: formation, process parameters, postprocessing, challenges, economic aspects, and future research directions},
  author={Haribaskar, R and Kumar, T Sampath},
  journal={3D Printing and Additive Manufacturing},
  volume={11},
  number={4},
  pages={1629--1655},
  year={2024},
  publisher={SAGE Publications Sage CA: Los Angeles, CA}
}

@article{poudel2022feature,
  title={Feature-based volumetric defect classification in metal additive manufacturing},
  author={Poudel, Arun and Yasin, Mohammad Salman and Ye, Jiafeng and Liu, Jia and Vinel, Aleksandr and Shao, Shuai and Shamsaei, Nima},
  journal={Nature Communications},
  volume={13},
  number={1},
  pages={6369},
  year={2022},
  publisher={Nature Publishing Group UK London}
}

@article{wang2026porosity,
  title={Porosity defects in additively manufactured metal materials: Formation mechanisms, impact on performance and regulation},
  author={Wang, Lei and Feng, Shengzhou and Wang, Yonggang and Zhao, Xiang and Ge, Jiaxing and Gao, Tianxi and Di, Fuqiang},
  journal={International Materials Reviews},
  volume={71},
  number={2},
  pages={97--128},
  year={2026},
  publisher={SAGE Publications Sage UK: London, England}
}

@article{zhan2025assessment,
  title={Assessment of the effect of the process-induced porosity defects on the fatigue properties of wire arc additive manufactured Al--Si--Mg alloy},
  author={Zhan, Teng and Xu, Ke and Fan, Zhipeng and Xiang, Hanlin and Xu, Congchang and Mei, Tianjiao and Wei, Yuanyuan and Chen, Wentao and Li, Luoxing},
  journal={Journal of Materials Research and Technology},
  volume={35},
  pages={777--791},
  year={2025},
  publisher={Elsevier}
}

@inproceedings{ouidadi2023defect,
  title={Defect segmentation from x-ray computed tomography of laser powder bed fusion parts: A comparative study among machine learning, deep learning, and statistical image thresholding methods},
  author={Ouidadi, Hasnaa and Xu, Boyang and Guo, Shenghan},
  booktitle={International Manufacturing Science and Engineering Conference},
  volume={87233},
  year={2023},
}

@article{ledwaba2025development,
  title={Development of AI crack segmentation models for additive manufacturing},
  author={Ledwaba, Tebogo and Steenkamp, Christine and Chmielewska-Wysocka, Agnieszka and Wysocki, Bartlomiej and du Plessis, Anton},
  journal={Tomography of Materials and Structures},
  volume={7},
  pages={100053},
  year={2025},
  publisher={Elsevier}
}

@inproceedings{lora_2022,
  title     = {{LoRA}: Low-Rank Adaptation of Large Language Models},
  author    = {Hu, Edward J. and Shen, Yelong and Wallis, Phillip and Allen-Zhu, Zeyuan and Li, Yuanzhi and Wang, Shean and Wang, Lu and Chen, Weizhu},
  booktitle = {International Conference on Learning Representations (ICLR)},
  year      = {2022}
}

@inproceedings{ptq4sam_2024,
  title={{PTQ4SAM}: Post-Training Quantization for Segment Anything},
  author={Lv, Chengtao and Chen, Hong and Guo, Jinyang and Ding, Yifu and Liu, Xianglong},
  booktitle={IEEE/CVF Conference on Computer Vision and Pattern Recognition},
  pages={15941--15951},
  year={2024}
}

@inproceedings{xctsam_2026,
  title={{XCT-SAM}: Sequential Parameter-Efficient Domain Adaptation of {SAM} for Industrial {XCT} Defect Segmentation},
  author={Hasan, Md Mahedi and Rahaman, Md Mushfiqur and Pachkovskiy, Alan and Ahmed, Imtiaz and Dawson, Jeremy and Das, Srinjoy},
  booktitle={IAPR Workshop on Machine Vision for Industrial Inspection (MVI2), International Conference on Pattern Recognition (ICPR)},
  year={2026}
}

@inproceedings{brecq_2021,
  title={{BRECQ}: Pushing the Limit of Post-Training Quantization by Block Reconstruction},
  author={Li, Yuhang and Gong, Ruihao and Tan, Xu and Yang, Yang and Hu, Peng and Zhang, Qi and Yu, Fengwei and Wang, Wei and Gu, Shi},
  booktitle={International Conference on Learning Representations},
  year={2021}
}

@inproceedings{qdrop_2022,
  title={{QDrop}: Randomly Dropping Quantization for Extremely Low-bit Post-Training Quantization},
  author={Wei, Xiuying and Gong, Ruihao and Li, Yuhang and Liu, Xianglong and Yu, Fengwei},
  booktitle={International Conference on Learning Representations},
  year={2022}
}

@inproceedings{adaround_2020,
  title={Up or Down? Adaptive Rounding for Post-Training Quantization},
  author={Nagel, Markus and Amjad, Rana Ali and van Baalen, Mart and Louizos, Christos and Blankevoort, Tijmen},
  booktitle={International Conference on Machine Learning},
  pages={7197--7206},
  year={2020}
}

@article{mobilesam_2023,
  title={Faster Segment Anything: Towards Lightweight {SAM} for Mobile Applications},
  author={Zhang, Chaoning and Han, Dongshen and Qiao, Yu and Kim, Jung Uk and Bae, Sung-Ho and Lee, Seungkyu and Hong, Choong Seon},
  journal={arXiv preprint arXiv:2306.14289},
  year={2023}
}

@inproceedings{efficientvit_sam_2024,
  title={{EfficientViT-SAM}: Accelerated Segment Anything Model Without Performance Loss},
  author={Zhang, Zhuoyang and Cai, Han and Han, Song},
  booktitle={IEEE/CVF Conference on Computer Vision and Pattern Recognition Workshops},
  year={2024}
}

@misc{qnn_vtcm_sharing,
  title        = {{HTP VTCM Sharing}, {Qualcomm AI Engine Direct SDK} Documentation},
  author       = {{Qualcomm Technologies, Inc.}},
  howpublished = {\url{https://docs.qualcomm.com/bundle/publicresource/topics/80-63442-10/htp_vtcm_sharing.html}},
  note         = {Document 80-63442-10. Accessed September 2026}
}

@misc{lam2024hexagon,
  title        = {Qualcomm's {Hexagon} {DSP}, and now, {NPU}},
  author       = {Chester Lam},
  howpublished = {Chips and Cheese, \url{https://chipsandcheese.com/p/qualcomms-hexagon-dsp-and-now-npu}},
  year         = {2024},
  note         = {Accessed September 2026}
}

@misc{qai_hub_docs,
  title        = {Qualcomm {AI} {Hub} Documentation},
  author       = {{Qualcomm Technologies, Inc.}},
  howpublished = {\url{https://app.aihub.qualcomm.com/docs/}},
  note         = {Accessed September 2026}
}

\clearpage
\appendix
\section*{Appendix}

\section{Training and evaluation details}
\label{app:train}
The synthetic benchmark consists of five held-out CycleGAN-generated splits,
\texttt{test1}, \texttt{test2}, \texttt{test4}, \texttt{test5} and \texttt{test6}, of $100$
slices each; \texttt{test4} and \texttt{test5} carry pore labels only. The real benchmark is
the NIST AM XCT set, specimens \texttt{test2} to \texttt{test6}, which has pore labels only.
Every checkpoint is scored with a fixed, per-split post-processing configuration (threshold,
morphological opening size, minimum blob size and, for one NIST specimen, intensity inversion
with non-zero valid-region masking) that was calibrated once and then applied identically to
every model, so that comparisons isolate the model rather than the threshold. Defect sizes
in these slices are small and widely spread within a class: over the $100$ slices of
\texttt{test1} at their native $768\times768$ resolution, connected pore components have a
median area of $14$ pixels and a $95$th percentile of $58$, with a median within-slice
maximum-to-minimum area ratio of $40$; inclusion components have a median of $22$ pixels and
a $95$th percentile of $71$, and there are about $2.3$ times fewer of them ($625$ against
$1{,}429$).

The validation split used for early stopping and best-checkpoint selection is the synthetic
\texttt{test6}, which is also one of the five reported splits; XCT-SAM used \texttt{test3}
for this purpose, which is not part of this dataset. The five-split means therefore include
the selection split, and the per-split post-processing settings were calibrated on the
respective split, including the NIST specimens, then applied identically to every model.
IoU is computed per image and averaged, with an image that has no defect and no predicted
pixel scored as $1$. This convention matters for inclusions, which are absent from most
slices: for the selected $(2,8)$ model at $1024^2$ the inclusion means over images are
$26.1$, $89.0$ and $31.3$ on \texttt{test1}, \texttt{test2} and \texttt{test6} ($48.9$
averaged), whereas over defect-bearing images alone they are $25.3$, $0.0$ and $10.2$ ($11.8$
averaged; \texttt{test2} has $89$ empty slices and the head predicts nothing on the other
$11$). For pores the two conventions give $36.1$ and $32.7$. The defect-only figures are the
ones to read for segmentation quality; the image-averaged figures are what
Table~\ref{tab:vitb} and the XCT-SAM comparison report.

Both heads are trained with AdamW at a base learning rate of $3\times10^{-4}$ with layer-wise
decay of $0.9$ per block toward the input, $10\%$ linear warmup followed by cosine decay,
weight decay $10^{-3}$, gradient clipping at $1.0$ and bfloat16 autocast, for a $30$-epoch
budget with early stopping on validation IoU. Inclusion-positive images are oversampled five
times by a weighted sampler, and augmentation is horizontal and vertical flips and $90^\circ$
rotations applied jointly to the image and both masks. The segmentation loss is Dice plus
focal with $\alpha_f=0.25$ and $\gamma_f=2$; a Focal--Tversky
alternative~\cite{focal_tversky_2019} with $(\alpha,\beta,\gamma)=(0.3,0.7,0.75)$, which
penalises missed defects more heavily than false alarms, is implemented but was not used for
the reported runs. The inclusion loss is averaged per image rather than over the flattened
batch, so that one densely labelled oversampled slice cannot dominate a step.

\paragraph{Training at $512^2$.} To test whether the accuracy lost at half resolution can be
recovered by training at that resolution, a ViT-H DCM-SAM with the default expert count was
trained from scratch at $512^2$ (position embedding resampled before training, $30$-epoch
budget, best checkpoint at epoch $5$, early-stopped at epoch $25$) and evaluated with the
$1024^2$ post-processing settings and, separately, with the blob-size filters rescaled by a
quarter for the smaller image. Table~\ref{tab:res512} compares it with its $1024^2$
counterpart of the same configuration and with the deployed ViT-B model. Retraining recovers
most of the pore loss seen zero-shot but not all of it: the pore head ends $5.6$ points below
its $1024^2$ twin on the image-averaged metric, and $12$ to $50\%$ lower per split on
defect-bearing images (\texttt{test4} $47.4$ to $40.2$, \texttt{test2} $33.0$ to $17.6$,
\texttt{test6} $14.2$ to $9.8$). Inclusion is unaffected or, once the filters are
rescaled, improves on two splits, which fits the size argument of \S\ref{sec:method}: pores
are the class that falls below what the model resolves at half resolution. The table also reads as a choice between the two ways of fitting the device, and the two
land on opposite sides of a defect-size boundary: dropping the resolution costs pores,
dropping the backbone costs inclusions. We took the side that keeps pores, the class that
drives certification and the only one the real data can score. The $512^2$ side rests on a
single run at one seed, was never evaluated on NIST, and was never put on the device with its
adapters.

\begin{table}[h]
\centering\footnotesize
\caption{Lowering the resolution against lowering the backbone. The same ViT-H configuration
trained and evaluated at $1024^2$ and at $512^2$ (``rescaled'' applies blob-size filters
scaled to $512^2$), and the deployed ViT-B $(2,8)$ model at $1024^2$. IoU in \%, mean over
the five synthetic splits; ``defect-only'' averages over defect-bearing images only.}
\label{tab:res512}
\begin{tabular}{lccc}
\toprule
 & ViT-H, $1024^2$ & ViT-H, $512^2$ (raw / rescaled) & ViT-B $(2,8)$, $1024^2$ \\
\midrule
Pore IoU, image-averaged & 37.3 & 31.7 / 30.8 & \textbf{36.1} \\
Pore IoU, defect-only & 35.8 & 27.9 / 27.6 & \textbf{32.7} \\
Inclusion IoU, image-averaged & 52.5 & \textbf{51.8} / 46.6 & 48.9 \\
Inclusion IoU, defect-only & 15.5 & 9.3 / \textbf{20.7} & 11.8 \\
Real NIST pore IoU & -- & -- & \textbf{64.2} \\
Runs on the IQ-9075 with adapters & no & -- & \textbf{yes, verified} \\
\bottomrule
\end{tabular}
\end{table}

\section{Seeding confound}
\label{app:seed}
An early version of the training pipeline drew the weighted-sampler and data-loader worker
seeds from the global random generator. The two heads are constructed sequentially before
training, and instantiating a Conv-LoRA head consumes a number of draws proportional to its
expert count, so varying $M_{\text{incl}}$ silently altered the data ordering seen by the
\emph{pore} head despite the heads sharing no parameters or gradients. Two runs with an
identical pore head differed by $4.9$ IoU points, larger than the effect being measured. The
sampler and worker seeds are now derived from dedicated generators seeded from the run seed,
independent of model construction; every result in the paper uses the corrected pipeline. We report it because nothing in the loss curves reveals it, and left in place it would have
inverted several of our conclusions.

\section{ViT-H expert-count study}
\label{app:vith}
\begin{table}[h]
\centering\small
\caption{ViT-H sweeps and corner runs. Each sweep row reports the swept head's IoU with the
fixed companion head's IoU from the same run; mean IoU (\%) over the five synthetic splits.}
\label{tab:vith}
\begin{tabular}{c cc cc}
\toprule
& \multicolumn{2}{c}{Pore sweep ($M_{\text{incl}}{=}8$)} & \multicolumn{2}{c}{Incl.\ sweep ($M_{\text{pore}}{=}8$)} \\
\cmidrule(lr){2-3}\cmidrule(lr){4-5}
$M$ & Pore & Incl. & Incl. & Pore \\
\midrule
1 & 35.5 & 53.8 & \textbf{54.3} & 34.5 \\
2 & \textbf{36.3} & 53.2 & 53.3 & \textbf{36.6} \\
4 & 36.2 & \textbf{54.3} & \textbf{54.4} & 32.6 \\
8 & 36.1 & 53.5 & 53.5 & 36.1 \\
\midrule
\multicolumn{5}{l}{Direct runs: $(1,1)$ $36.3$ / $53.6$;\quad $(2,1)$ \emph{(selected)} $34.6$ / $53.3$} \\
\bottomrule
\end{tabular}
\end{table}
In Table~\ref{tab:vith} the ViT-H inclusion sweep spans $1.1$ points across an $8\times$
change in mixture size and the pore sweep $0.8$ points above $M=1$: on this backbone the mixture size is close to
irrelevant once $M\ge2$. The ViT-B round (Table~\ref{tab:vitb}) behaves quite differently, degrading sharply at
$(8,8)$. It is worth adding why the per-head optima fail to compose there, with $(2,2)$ worse
than $(2,8)$ on both metrics. The two heads share no trainable parameters and no gradients,
so they cannot be interfering during optimisation; what changes is the data ordering each
head sees as the other head's expert count changes, the same mechanism as the seeding
confound of Appendix~\ref{app:seed}.

\section{Trainable parameter budget}
\label{app:params}
\begin{table}[h]
\centering\small
\caption{Trainable parameters per defect head, counted from the instantiated model ($r=2$,
$M=8$). A checkpoint contains exactly these. The mask decoder is the same module on both
backbones.}
\label{tab:params}
\begin{tabular}{lrrrr}
\toprule
Component & ViT-H & Share & ViT-B & Share \\
\midrule
Frozen SAM backbone (encoder, prompt encoder) & 637.0\,M & frozen & 89.7\,M & frozen \\
Conv-LoRA $A,B$ & 327{,}680 & 0.051\% & 73{,}728 & 0.079\% \\
Convolutional experts & 9{,}728 & 0.002\% & 3{,}648 & 0.004\% \\
MoE gating ($W_g,W_n$) & 1{,}024 & $<$0.001\% & 384 & $<$0.001\% \\
Mask decoder (incl.\ IoU head) & 4.06\,M & 0.633\% & 4.06\,M & 4.33\% \\
\midrule
\textbf{Trainable per head} & \textbf{4.40\,M} & \textbf{0.69\%} & \textbf{4.13\,M} & \textbf{4.41\%} \\
Both heads & 8.79\,M & 1.36\% & 8.27\,M & 8.44\% \\
\bottomrule
\end{tabular}
\end{table}
Table~\ref{tab:params} breaks the budget down by component. On ViT-B the adapter accounts
for $0.08\%$ of the model, and the mask decoder for $98\%$ of what is trained. None of the adapter can be folded back into the frozen weights, since its
expert branch contains a GELU, convolutions and a gate that depends on the input, so the
adapters are executed at inference like any other layer.

\section{Why the adapted encoder did not fit, and what fixed it}
\label{app:attn}
The adapted ViT-B encoder exported as trained needs $4.07$\,GB at $1024^2$ and is refused
by the QNN graph serializer, which reports ``graph requires estimated allocation of 4272953
KB, limit is 3670016 KB; error during serialize: memory usage too large'' when the graph is
compiled through Qualcomm AI Hub~\cite{qai_hub_docs} (compile job \texttt{jpyxke3l5}); that
$3.76$\,GB limit is target-dependent and is not a published hardware figure. The stock ViT-B
encoder runs. The adapter holds $0.09\%$ of the encoder's parameters, and on a GPU bypassing it leaves
peak memory unchanged, so parameters
and dynamic memory do not explain the gap. For scale, at $1024^2$ in FP16 the attention
matrix is $384$\,MB for ViT-B's $12$ heads and $512$\,MB for the $16$ heads of ViT-L and
ViT-H, the $q,k,v$ and MLP activations are $18$ to $40$\,MB, and the weights are $171$, $588$
and $1{,}215$\,MB; the $8$\,MB on-chip scratchpad is exceeded $48$ to $64$ times by every
backbone's attention matrix, so it is tiled through and does not discriminate between them.
The difference lay in two code paths. We had exported the stock encoder through
HuggingFace's scaled-dot-product (SDPA) attention, while the vendored Conv-LoRA code inherits
the older eager attention, which adds the decomposed relative-position terms as two chained
broadcasts and then takes an FP32 softmax. To separate the two effects we compiled the
variants in Table~\ref{tab:attn} with identical options and ran them on the device at
$1024^2$. The stock encoder runs with eager attention, so the eager path alone is not the
cause; the adapted encoder runs once its attention is rewritten in SDPA form, so the adapter
alone is not the cause either. Only the two together exceed the ceiling. The rewrite is the same computation: encoder outputs
agree to $1.2\times10^{-5}$ on ViT-B and the final masks differ in $0$ of $1{,}048{,}576$
pixels. The SDPA form is now the default, with the eager path kept behind a flag.

\begin{table}[h]
\centering\small
\caption{Attention-formulation diagnostic on the IQ-9075 at $1024^2$, FP16. Cosine is against
the corresponding PyTorch FP32 output on a real XCT slice.}
\label{tab:attn}
\begin{tabular}{llllc}
\toprule
Graph & Conv-LoRA & Attention & On device & Cosine \\
\midrule
Stock ViT-B                     & no  & SDPA  & runs & 0.99994 \\
Stock ViT-B                     & no  & eager & runs & 0.99994 \\
DCM-SAM pore head, as trained   & yes & eager & fails, needs 4.07\,GB & n/a \\
DCM-SAM pore head, rewritten    & yes & SDPA  & \textbf{runs} & 0.99988 \\
DCM-SAM inclusion head, rewritten & yes & SDPA & \textbf{runs} & 0.99995 \\
\midrule
\multicolumn{5}{l}{Static one-hot routing vs.\ dynamic top-$1$ (PyTorch, eight slices): $13$ of $8.4$M mask pixels differ} \\
\multicolumn{5}{l}{End to end on device, one slice per head: $96$ (pore) and $0$ (inclusion) of $1{,}048{,}576$ mask pixels differ} \\
\bottomrule
\end{tabular}
\end{table}

\section{NPU graph construction details}
\label{app:npu}
Beyond the attention constraint of \S\ref{sec:exp:backbone}, two further operator-level
constraints determine the mixed-precision INT8 scheme used for stock ViT-B. Exact GELU emits
an $\mathrm{erf}$ operator for which the converter has no layout inference; the tanh
approximation is substituted, with maximum output deviation $1.0\times10^{-3}$ (ViT-B) and
$3.5\times10^{-3}$ (ViT-L). A partially quantized normalisation layer presents mixed
datatypes and is rejected, so quantization is restricted to weight-bearing operators
(convolutions and weight--activation matmuls), leaving normalisation, softmax and elementwise
operators in floating point; the excluded operators hold almost no parameters. For DCM-SAM
the static gate required three attempts: \texttt{ScatterElements} is emitted at an opset the
converter does not accept, the backend has no validated datatype combination for
\texttt{OneHot}, and the working form is an equality between $\arg\max$ and
$\mathrm{arange}(M)$. Serialising the decoder without the encoder weights reduced that
artifact from $358$\,MB to $15.7$\,MB. The measured $4.1\times$ slowdown from $M=2$ to $M=8$
is well below what the experts alone would predict: expert $i$ convolves a map upsampled by
$i$, so expert compute grows as $\sum_{i=1}^{M} i^2$, from $5$ to $204$, a factor of $40$;
the rest of the encoder, which does not change with $M$, dilutes that to the $4.1\times$
observed, which is also why the scaling is not linear in $M$. Retargeting SAM to an input size $S'$ other than
$1024$ takes three changes: relax the size check in the patch embedding, resample the learned
position embedding bicubically from the $64\times64$ grid to $S'/16$, and update the
patch-grid size held by the prompt encoder, which sets the shape of the dense ``no prompt''
embedding that the decoder adds to the image embedding. Without the last step the decoder
receives tensors of the wrong shape; the relative-position tables need nothing, since they
are interpolated to the query and key extent at every call.

\section{Limitations and next steps}
\label{app:limits}

\begin{table}[t]
\centering\scriptsize
\renewcommand{\arraystretch}{0.92}
\caption{Measured cost of the four graphs on the IQ-9075, FP16. The model is deployed at
$1024^2$ (\S\ref{sec:exp:deploy}); the per-operator profiler attaches at $512^2$, so the
encoder rows are timed there, from an export of the same weights. At $1024^2$ the decoders
measure $15.4$ and $15.3$\,ms. Peak memory excludes the loaded binary.}
\label{tab:deploy}
\begin{tabular}{llrrrr}
\toprule
Artifact & Experts & NPU ops & Inference (ms) & Peak mem.\ (MB) & Load (ms) \\
\midrule
Pore encoder      & $M=2$ & 1012 & 1980.8 & 11.8 & 825 \\
Pore decoder      & n/a   &  205 &    3.8 &  9.6 & 160 \\
Inclusion encoder & $M=8$ & 1444 & 8134.3 & 13.9 & 868 \\
Inclusion decoder & n/a   &  205 &    3.8 &  9.6 & 167 \\
\bottomrule
\end{tabular}
\vspace{-3mm}
\end{table}
Several caveats bear on how far these numbers should be pushed. Every configuration in
Table~\ref{tab:vitb} was trained once, and we have already seen a
$4.9$-point swing from a seeding confound on this backbone (Appendix~\ref{app:seed}), so
differences under about three points are not resolved; the choice of $(2,8)$ was made on the synthetic benchmark, and on NIST it is the lowest of
the nine, though only by $2.1$ points, which we do not read as a real difference either. The baseline numbers in
Table~\ref{tab:baselines} are taken from XCT-SAM rather than re-run under our protocol, and
they are ViT-H models while ours is ViT-B, so that table is indicative until a
protocol-matched re-evaluation is done. The deployment is FP16: INT8 was blocked at the
operator level for the attention products (\S\ref{sec:exp:backbone}) and no integer-quantized accuracy is reported anywhere in the paper. On-device numerical equivalence was established on one real XCT slice per
head at $1024^2$ and on the encoder outputs of the stock backbone; the per-op profiler cannot
attach to a $1024^2$ encoder graph on this device, so encoder latency at the deployed
resolution is not measured and Table~\ref{tab:deploy} reports $512^2$. Resolution cannot be traded for speed even with retraining: the $512^2$ evidence is one
ViT-H run, one seed (Appendix~\ref{app:train}). Finally, because
the adapters sit inside the encoder, the encoder is executed once per head, and each encoder
binary carries its own copy of the frozen weights, so sharing the backbone saves storage
during training but not on the device. Repeated seeds, a protocol-matched baseline
comparison, an internal shared-adapter ablation, and on-board timing of the $1024^2$
binaries are the next steps.

\section{Per-split NIST results}
\label{app:nist}
\begin{table}[h]
\centering\small
\caption{Per-split NIST results for the selected $(2,8)$ ViT-B model. Precision and recall
at the $5$\,px tolerance used for Tol-F1. $^{*}$\texttt{test6} uses inversion and non-zero
valid-region masking, calibrated for this specimen's highly porous, inverted-intensity
appearance.}
\label{tab:nist-split}
\begin{tabular}{lccccc}
\toprule
Split & IoU (\%) & Dice (\%) & Tol-F1 (\%) & Precision (\%) & Recall (\%) \\
\midrule
NIST \texttt{test2}   & 50.0 & 66.5 & 85.1 & 74.2 & 100.0 \\
NIST \texttt{test3}   & 64.2 & 78.0 & 94.1 & 99.2 & 90.0 \\
NIST \texttt{test4}   & 54.5 & 70.4 & 89.0 & 80.6 & 99.9 \\
NIST \texttt{test5}   & 66.6 & 79.9 & 96.2 & 98.3 & 94.2 \\
NIST \texttt{test6}$^{*}$ & 85.4 & 92.0 & 96.8 & 93.8 & 100.0 \\
\midrule
\textbf{Mean}    & \textbf{64.2} & \textbf{77.4} & \textbf{92.2} & \textbf{89.2} & \textbf{96.8} \\
\bottomrule
\end{tabular}
\end{table}
In Table~\ref{tab:nist-split}, \texttt{test2} and \texttt{test4} combine near-perfect recall
with markedly lower precision: the pore head over-predicts on these specimens, which depresses IoU while leaving Tol-F1 at
$85$--$89\%$.

\end{document}